\documentclass[10pt,twocolumn,letterpaper]{article}

\usepackage{bbding}
\usepackage{iccv}
\usepackage{times}
\usepackage{epsfig}
\usepackage{graphicx}
\usepackage{amsmath}
\usepackage{amssymb}
\usepackage{booktabs}
\usepackage{multirow}
\usepackage{color}
\usepackage[misc]{ifsym}
\usepackage{bm}
\usepackage{mathrsfs}
\usepackage{amsfonts}

\DeclareRobustCommand*{\IEEEauthorrefmark}[1]{%
	\raisebox{0pt}[0pt][0pt]{\textsuperscript{\footnotesize\ensuremath{#1}}}}

\usepackage[breaklinks=true,bookmarks=false]{hyperref}

\iccvfinalcopy 

\def\iccvPaperID{5714} 

\ificcvfinal\fi

\def\ie{\emph{i.e}\onedot} 
 
\def\etc{\emph{etc}\onedot} 
 
\def\etal{\emph{et al}\onedot}

\begin{document}

\title{Multi-Source Domain Adaptation for Object Detection}

\author{Xingxu Yao\IEEEauthorrefmark{1}, Sicheng Zhao\IEEEauthorrefmark{2}, Pengfei Xu\IEEEauthorrefmark{3}, Jufeng Yang\IEEEauthorrefmark{1}	\\
\IEEEauthorrefmark{1}Nankai University, China
\IEEEauthorrefmark{2}Columbia University, USA
\IEEEauthorrefmark{3}Didi Chuxing, China\\
{\tt\small yxx\_hbgd@163.com, schzhao@gmail.com, xupengfeipf@didiglobal.com, yangjufeng@nankai.edu.cn}
}


\maketitle
\ificcvfinal\thispagestyle{empty}\fi

\begin{abstract}
	To reduce annotation labor associated with object detection, an increasing number of studies focus on transferring the learned knowledge from a labeled source domain to another unlabeled target domain.
	However, existing methods assume that the labeled data are sampled from a single source domain, which ignores a more generalized scenario, where labeled data are from multiple source domains.
	For the more challenging task, we propose a unified Faster R-CNN based framework, termed Divide-and-Merge Spindle Network (DMSN), which can simultaneously enhance domain invariance and preserve discriminative power.
	Specifically, the framework contains multiple source subnets and a pseudo target subnet.
	First, we propose a hierarchical feature alignment strategy to conduct strong and weak alignments for low- and high-level features, respectively, considering their different effects for object detection.
	Second, we develop a novel pseudo subnet learning algorithm to approximate optimal parameters of pseudo target subset by weighted combination of parameters in different source subnets.
	Finally, a consistency regularization for region proposal network is proposed to facilitate each subnet to learn more abstract invariances.
	Extensive experiments on different adaptation scenarios demonstrate the effectiveness of the proposed model.

\end{abstract}

\section{Introduction}

As a fundamental task in computer vision, object detection has drawn much attention in the past decade~\cite{redmon2018yolov3,redmon2016you,liu2016ssd}.
With the development of convolutional neural networks (CNNs), some modern CNN-based detectors like Faster R-CNN~\cite{ren2015faster} have emerged and been successfully applied to many tasks, such as autonomous driving~\cite{geiger2012we, shan2019pixel}, face and pedestrian detection~\cite{jiang2017face, mao2017can}, \etc.
However, the high-quality performance of detectors is based on large-scale training images with annotated bounding boxes.
In the real world, variances exist between training and test images in many aspects, including object appearance, background, even taken time.
Due to these domain discrepancies, the performance on the test images may decrease dramatically.
Although annotating more training data of the new domain is able to alleviate the phenomenon, it is not an optimal strategy due to large time and labor costs.

\begin{figure}[t]
	\begin{center}
		\includegraphics[width=0.9\linewidth]{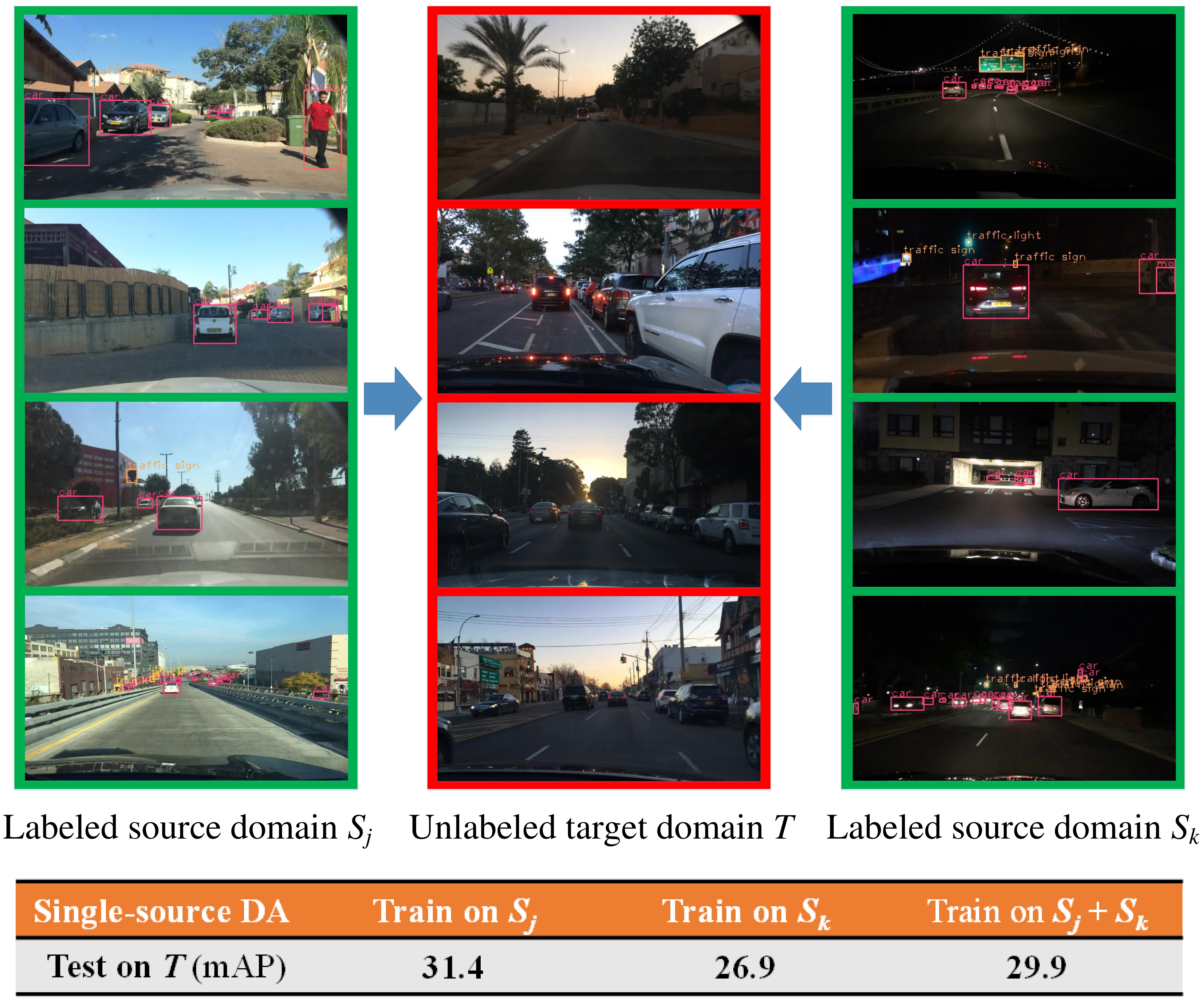}
		\caption{An example of domain shift in the multi-source scenario for object detection.
			The above images are sampled from different subsets of BDD100k~\cite{yu2018bdd100k}.
			Source domain $S_j$ and $S_k$ are taken in \textit{daytime} and \textit{night}, respectively, while the images in target domain $T$ are taken in \textit{dawn/dusk}.
			As shown in the above results, directly combining images from multiple sources and conducting single-source domain adaptation (DA) will cause performance decay compared with only using the best single source domain.
			Note that mAP denotes mean average precision.}
		\label{fig: abstract}
	\end{center}
\end{figure}

To mitigate the domain gap, unsupervised domain adaptation (UDA) has been widely used for object detection~\cite{chen2018domain, kim2019diversify, li2020deep, kim2019self}.	
Domain adaptive Faster R-CNN~\cite{chen2018domain} is a milestone study developed for tackling the domain shift problem in object detection.
In the work, image-level features and instance-level features are respectively aligned through an adversarial manner.
Following~\cite{chen2018domain}, a series of Faster R-CNN based adaptation models~\cite{zhuang2020ifan,zhu2019adapting} have emerged recently.
Considering that image-level alignment transfers massive unnecessary information background for object detection, Zhu~\etal~\cite{zhu2019adapting} and Saito~\etal~\cite{saito2019strong} pay more attention to align informative local regions.
To minimize the domain distribution disparity on each block, He~\etal~\cite{he2019multi} propose multi-adversarial Faster R-CNN to conduct layer-wise domain feature alignment.
%
%
However, existing algorithms for domain adaptive object detection assume that the source data are sampled from a single domain, which limits the generalization of the model.
We consider a more practical scenario that the source data are collected from multiple domains with different distributions.
As shown in Figure~\ref{fig: abstract}, directly combining two sources and performing single-source domain adaptation (DA) cause performance decay compared with the best result of a single domain, \ie 29.9\% \textit{vs.} 31.4\%.
It is mainly resulted from the mutual interference between different sources that exist serious domain discrepancy, which is also demonstrated in other tasks~\cite{zhao2019multi,peng2019moment,riemer2018learning}.
Therefore, we need to design a specialized framework for domain adaptive object detection from multiple sources.

Though multi-source domain adaptation has been explored for other tasks, such as image classification~\cite{zhao2019multi,xu2018deep} and segmentation~\cite{zhao2019multi_seg}, they all belong to the straightforward classification task regardless of image-level or pixel-level.
However, detection model like Faster R-CNN is a complex system for both regression and classification, and it contains multiple components including feature extractor, region proposal network (RPN), \etc.
In this paper, for the new task, we develop a framework termed Divide-and-Merge Spindle Network (DMSN), in which multiple source subnets and a pseudo subnet are included.
First, we propose a hierarchical feature alignment strategy.
Since low-level features have high resolution that is important to localization~\cite{qin2019thundernet}, we perform strong alignment for them of different domains.
For the high-level features that are important for object recognition, we weakly align each source and target in corresponding supervised source subnet.
%
Second, to approximate optimal parameters for the target domain, we develop a pseudo subnet learning (PSL) algorithm, in which the pseudo subnet is updated via exponential moving average (EMA)~\cite{tarvainen2017mean} parameters of different source subnets.
Finally, a new consistency regularization on region proposals are conducted between each source subnet and the pseudo target, which enables the model to capture more robust instance-level invariances for object detection.
%
In the testing phase, the predictions of each subnet are merged into final inference.
%
%
%
%

In summary, our contributions are threefold.
\begin{itemize}
	\item We propose to conduct domain adaptation for object detection from multiple sources.
	To the best of our knowledge, this is the first work
	on multi-source domain adaptation for object detection that contains both classification and regression.
	%
	%
	\item We propose a unified framework termed DMSN to tackle the new problem.
	The characteristics of each source domain are preserved in each independent supervised source subnet.
	Meanwhile, an optimal pseudo subnet is approximated by aggregating parameters of different source subnets.
	%
	%
	%
	\item A new consistency regularization is designed to facilitate each source subnet to propose similar region proposals with the pseudo subnet, so that the model is able to learn more abstract invariances.
	The results of extensive experiments demonstrate the effectiveness of our proposed framework.
\end{itemize}
%
\section{Related Work}
\subsection{Object Detection}
Object detection~\cite{ren2015faster,girshick2014rich} is a fundamental task in computer vision, and it has received booming development during the last decade.
The methods that are based on the deep convolutional neural network (CNNs) can be classified into two types: one-stage model~\cite{liu2016ssd, redmon2016you, redmon2017yolo9000} and two-stage model~\cite{ren2015faster}.
Two-stage model employs region proposal-based detectors, in which a mass of region proposals are generated in advance for refining and revision.
Originally, Fast R-CNN~\cite{girshick2015fast} utilizes selective search to obtain candidate bounding boxes.
Later in Faster R-CNN~\cite{ren2015faster}, region proposals are learned by the proposed RPN, which obviously saves computing consumptions.
Recently, some one-stage models are introduced to accelerate the speed of detection.
These methods apply predefined anchor boxes or points near the center of objects to predict bounding boxes.
In this paper, we use Faster R-CNN as our base detector following~\cite{shen2019scl, zhu2019adapting, hsu2020progressive}, and transfer knowledge from multiple source domains to the new target domain.

\begin{figure*}[t]
	\begin{center}
		\includegraphics[width=0.95\linewidth]{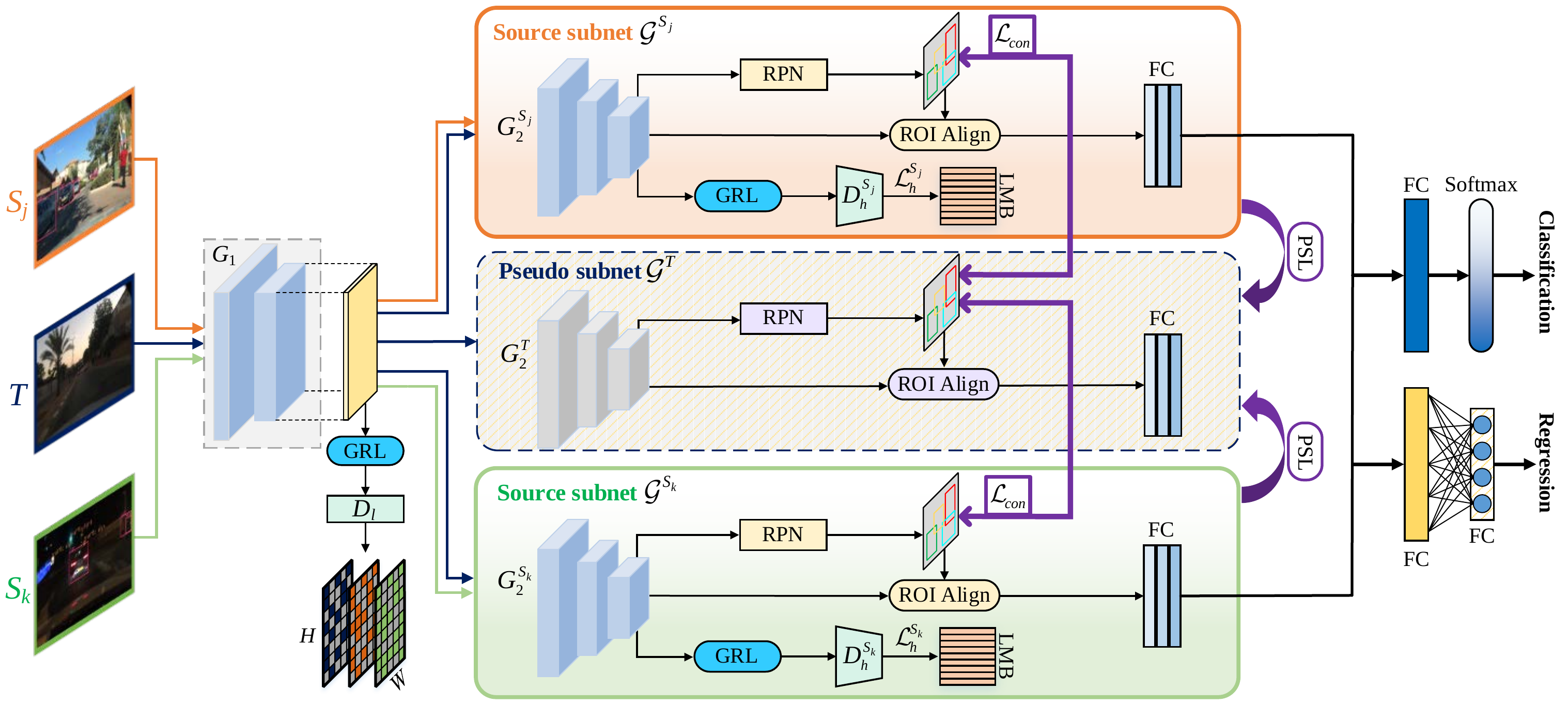}
		\caption{Pipeline of the proposed DMSN. The feature extractor is divided into two parts, including $G_1$ and $G_2$, where $G_2 = \{ \{G^{S_i}_2 \}^M_{i=1}, G^T_2\}$. In the first training phase, features from $G_1$ are aligned through GRL and domain discriminator $D_l$ trained by least-squares loss. Then the images of domain $S_j$ and $T$ are fed into the source subnet $\mathcal{G}^{S_i}$, $i=1,... , M$ to perform the high-level source-target domain alignment and source domain detection learning. In the second training phase, we maintain the learning objective of the first phase and add consistency regularization ($\mathcal{L}_{con}$) for RPN and pseudo subnet learning (PSL). LMB denotes loss memory bank. Note that the arrows or lines in \textit{\textbf{{\textcolor[RGB]{128,0,128}{purple}}}} denote the operation in the second training phase.
		}
		\label{fig: pipline}
	\end{center}
\end{figure*}

\subsection{Domain Adaptation for Object Detection}
Due to the labor consumption of data annotation and domain shift existing in different domains, domain adaptive object detection~\cite{xu2020cross,arruda2019cross,wang2019few,wang2019towards} has emerged to solve these problems.
Domain adaptive Faster R-CNN~\cite{chen2018domain} is a milestone two-stage object detector to diminish the domain gap between different datasets.
The work is based on Faster R-CNN~\cite{ren2015faster} and aligns the distribution on image-level and instance-level by domain classifier by adversarial learning.
Besides, the consistency between image-level and instance-level alignment is enforced to learn a more robust model.
Along this idea, a series of frameworks~\cite{cai2019exploring, kim2019diversify, khodabandeh2019robust} are developed for domain adaptive object detection.
For example, He~\etal~\cite{he2019multi} propose to conduct feature alignment in multiple layers, while Saito~\etal~\cite{saito2019strong} exploit strong and weak adaptation in local and strong features, respectively.
Apart from aligning different domains on feature-level, Hsu~\etal~\cite{hsu2020progressive} also employ CycleGAN~\cite{hsu2020progressive} to map source images into an intermediate domain through synthesizing the target distributions on the pixel-level.
%
%
When directly applying to adaptive object detection with multiple sources, a more practical scenario, most of these methods suffer from low performances.
In this paper, we develop a specialized network to handle source images from multiple domains.

\subsection{Multi-Source Domain Adaptation}
Multi-source domain adaptation (MSDA)~\cite{sun2015survey, zhao2020multi, takshit20, yangbls20, li2020eccv_oml} assumes that the training instances are from multiple domains.
Recently, several studies follow the routine of single-source DA to utilize a shared network to forcedly align the multiple source distributions.
For instance, multi-source domain adversarial network (MDAN)~\cite{zhao2018adversarial} trains a discriminator to distinguish the source and target data, while the feature extractor aims to obtain domain-invariant features by adversarial learning.
To better diminish the discrepancy between domains, Peng~\etal~\cite{peng2019moment} propose to transfer the knowledge from multiple sources to the target by aligning moments of their feature distributions.
However, these methods may result in performance decay due to the loss of the discriminative power of each domain.

Besides, there is another routine based on the theoretical analysis~\cite{ben2010theory, blitzer2008learning, mansour2009domain} for MSDA, which demonstrates that the target distribution can be approximated by a mixture of the multiple source distributions.
To calculate more accurate combination weights for source domains, Hoffman~\etal~\cite{hoffman2018algorithms} develop a novel bound using DC-programming.
Therefore, weighted combination of multiple source domains has been widely used for MSDA~\cite{wang2020learning}.
Xu~\etal~\cite{xu2018deep} develop a deep cocktail network (DCTN) to train a classifier for each source domain and combine the predictions of different classifiers weighted by perplexity scores.
In~\cite{zhao2019multi}, Zhao~\etal first fix the specialized feature extractor of
each source and adversarially map the target data into the feature space of each source.
Then, the prediction is generated by aggregating different source predictions weighted by the discrepancy between each source and the target.
Different from them, we introduce a pseudo subnet for the target domain, which is updated by directly weighting parameters of each source subnet.
Also, existing MSDA algorithms~\cite{zhao2019multi,li2018extracting,xu2018deep} forcibly align different domains without considering the loss of the feature discriminativity.
In this paper, we can not only learn more domain invariance but also preserve discriminative property of features.
Besides, we are the first to focus on MSDA for object detection.
%

\section{Method}
\subsection{Overview}
In unsupervised MSDA, we consider the scenario that there are $M$ labeled source domains $S_1, S_2, ..., S_M$ and one unlabeled target domain $T$.
In the $i$-th source domain $S_i = \{(x_i^j, B_i^j)\}_{j=1}^{N_i}$, suppose $x_i^j$ represents the $j$-th images and $B_i^j$ denotes the corresponding bounding box annotation.
Note that $N_i$ is the number of images in the $i$-th source domain.
In the unlabeled target domain $T = \{x_T^j\}_{j=1}^{N_T}$, the $j$-th image is represented by $x_T^j$, and $N_T$ denotes the images of target domain.
In this problem, our goal is to learn an object detector that can correctly detect the object in images from the target domain based on multiple labeled source domains $\{S_i\}_{i=1}^{M}$ and unlabeled training data in the target domain $T$.
We emphasize the following two hypotheses in the paper:
(1) homogeneity, \ie, $x_i^j \in \mathbb{R}^{d}, x_T^j \in \mathbb{R}^{d}$, indicating that the feature of data from all the domains are in identical space with the same dimension but different distributions;
(2) closed set, meaning that the data from all the domains share the same category space.

We propose a novel framework termed Divide-and-Merge Spindle (DMSN) and the pipeline is shown in Figure~\ref{fig: pipline}.
Our learning process is divided into two phases.
In the first training phase, we strongly align low-level features for all the domains, while the high-level features are weakly aligned between each source domain and the target domain.
Meanwhile, we also train each labeled source subnet $\mathcal{G}^{S_i}$ separately supervised by instance-level annotations, which enables it to effectively learn the characteristics of each source.
In the second training phase, the learning objective in the first training phase is maintained.
%
%
We conduct a consistency learning algorithm on region proposals.
Besides, we propose a dynamic weighting (DW) strategy to aggregate the parameters of other source subnets for updating the parameters of the pseudo subnet.

\subsection{Hierarchical Feature Alignment}
\label{Sec3.2}
As shown in Figure~\ref{fig: pipline}, the backbone is divided into two parts, \ie $G_1$ and $G_2 =\{ \{G^{S_i}_2 \}^M_{i=1}, G^T_2\} $.
We perform different alignment strategies for the features of different levels.
At the same time, the supervised learning for object detection for each labeled source domain is conducted in the corresponding source subnet.

\noindent \textbf{Low-level feature alignment.} Considering the fact that low-level features are scarcely associated with high-level semantics, we aim to learn domain-invariant features with shared parameters.
Furthermore, with the high resolution, low-level features benefit to improve the localization ability~\cite{qin2019thundernet,chen2020refinedetlite}.
Therefore, we strongly conduct align local features in lower layers using a least-squares loss inspired by~\cite{saito2019strong, zhu2017unpaired} to train the domain discriminator $D_l$, consisting of 1$\times$1 convolutional layers.
A gradient reversal layer (GRL)~\cite{ganin2015unsupervised} is placed between the backbone and domain discriminator $D_l$ to implement adversarial learning.
The low-level feature maps extracted from $G_1$ are input into $D_l$, and then the dimension of the channel is reduced to $M+1$, representing $M$ source domains and one target domain.
The width and height of the feature maps are denoted with $W$ and $H$, respectively.
%
%
We use $D_l(G_1(x))^k_{wh}$ to represent each location in $k$-th channel-wise output for image $x$, where $k= 1, 2, ..., M+1$.
Different from~\cite{saito2019strong}, for images $\{(x_i^j)\}_{j=1}^{N_i}$ of $i$-th source domain, we aim to not only align the features with the target domain but also bridge the gap between source domains:
\begin{small} 
\begin{equation}
\begin{aligned}
\mathcal{L}^{S_i}_l &= \frac{1}{N_iWH}\sum_{j=1}^{N_i}  \sum_{w=1}^{W} \sum_{h=1}^H  \bigg(\big(1-D_l(G_1(x_i^j))^i_{wh}\big)^2 \\
&+\sum_{k\neq i}^{M+1}  \big(D^k_l\big(G_1(x_i^j)\big)^k_{wh}\big)^2\bigg).
\end{aligned}
\end{equation}%
\end{small}

For target images $\{x_T^j\}_{j=1}^{N_T}$, the least-squares loss is formulized as:
\begin{footnotesize}
\begin{equation}
\begin{aligned}
\mathcal{L}^T_l &= \frac{1}{N_TWH}\sum_{j=1}^{N_T}  \sum_{w=1}^{W} \sum_{h=1}^H  \big(1-D_l(G_1(x_T^j))^{M+1}_{wh}\big)^2.
\end{aligned}
\end{equation}%
\end{footnotesize}
Since the two losses are conducted on each location of the feature maps, the local features can be well aligned.
The overall loss on low-level features is formulized as:
\begin{small}
\begin{equation}
\begin{aligned}
\mathcal{L}_l &= \sum_{i=1}^M \mathcal{L}_l^{S_i}+\mathcal{L}_l^T.
\end{aligned}
\end{equation}%
\end{small}

\noindent \textbf{High-level feature alignment.}
In the higher layers, the network is divided into $M+1$ branches, including $M$ source subnet $\{\mathcal{G}^{S_i}\}_{i=1}^{M}$ and a pseudo target subnet $\mathcal{G}^{T}$.
The source instances are fed into one of the corresponding specialized subnets, while the target instances are fed into all the subnets.
We define $G_2^{S_i}$ as the second-stage feature extractor for the $i$-th source domain, while $G_2^{T}$ denotes the second-stage feature extractor in the pseudo subnet.
In the source subnet $\mathcal{G}^{S_i}$,
we suppose that the high-level holistic feature of $j$-th image from the $i$-th source and target are respectively represented as $F_{S_{ij}}$ and $F_{T_{ij}}$, respectively, where $F_{S_{ij}} = G_2^{S_i}(G_1(x_i^j))$ and $F_{T_{ij}} = G_2^{S_i}(G_1(x_T^j))$.
To preserve characteristics of each source, we aim to weakly align distributions of holistic features for each $G_2^{S_i}$ via GRL and source-target domain discriminator $D^{S_i}_h$.
Inspired by~\cite{saito2019strong}, we employ focal loss~\cite{lin2017focal} to train $D^{S_i}_h$ of $G_2^{S_i}$ as follows:
\begin{small}
\begin{equation}
\begin{aligned}
\label{gamma}
\mathcal{L}^{S_i}_h &= \frac{1}{N_i}\sum_{j=1}^{N_i}\big(1-D^{S_i}_h\big(F_{S_{ij}})\big)^\gamma\log\big(D^{S_i}_h(F_{S_{ij}}) \big)\\ &+ \frac{1}{N_T}\sum_{j=1}^{N_T}D^{S_i}_h\big(F_{T_{ij}})^\gamma\log\big(1-D^{S_i}_h\big(F_{T_{ij}}) \big),
\end{aligned}
\end{equation}%
\end{small}
where $\mathcal{L}^{S_i}_h$ denotes the domain classifier loss on high-level features between the $i$-th source and the target.
$\gamma$ is used to controls the weight on hard examples.
The overall source-target domain discriminator loss is written as $\mathcal{L}_h = \sum^M_{i=1} \mathcal{L}^{S_i}_h.$

\subsection{Pseudo Subnet Learning}
After the first training phase, each source subnet has learned characteristics of corresponding source domain.
According to the theoretical proof in~\cite{ben2010theory, blitzer2008learning, mansour2009domain}, the target distribution can be approximated by the weighted combination of the distribution of different sources.
Existing studies on MSDA~\cite{xu2018deep,zhao2019multi} mainly focus on weighting different predictions in the inference phase.
In this paper, from a new perspective, we aim to directly learn a pseudo subnet $\mathcal{G}^T$ for target by weighting the parameters of each source subnet $\mathcal{G}^{S_i}$.
Inspired by mean teacher method~\cite{cai2019exploring, tarvainen2017mean,french2017self} used in semi-supervised learning, we obtain the pseudo subnet by exponential moving average (EMA) parameters of source subnets.
Different from~\cite{cai2019exploring}, which only averages parameters on the temporal sequence, we also consider how to aggregate the parameters of different source subnets.
To reasonably integrate the knowledge of each source, we design a dynamic weighting (DW) strategy based on the discrepancy between each source and target to conduct greater weight for more relevant sources.

As mentioned in Section~\ref{Sec3.2}, source-target classifier $D_h^{S_i}$ in $\mathcal{G}^{S_i}$ is employed to distinguish target images and source images of the $i$-th domain.
The larger value of $\mathcal{L}_h^{S_i}$ indicates that the distributions of the $i$-th source and the target are more similar.
Based on the important cues, we design a dynamic loss memory bank (LMB) to store the loss values of the past training steps.
We use $\mathcal{U}^t_i$ to denote LMB of the $i$-th source domain in $t$-th training step, \ie $\mathcal{U}^t_i = [\mathcal{L}^{S_i}_{h}(t-c+1),\mathcal{L}^{S_i}_{h}(t-c+2),... , \mathcal{L}^{S_i}_h(t)]$, where $\mathcal{L}^{S_i}_h(t)$ denotes the loss value of $t$-th training step and $c$ represents the length of $\mathcal{U}^t_i$.
The mean value of all the elements in $\mathcal{U}^t_i$ is denoted as $\mathcal{V}^t_{i}$.
The larger the value of $\mathcal{V}_{i}^t$ is, the more similar the distributions of the $i$-th source and the target are, and vice versa.
Therefore, we measure the discrepancy of $S_i$ and $T$ based on $\mathcal{V}^t_{i}$ at iteration $t$.
Considering LMB of all the sources, we define the relative similarity of the $i$-th source and the target at iteration $t$ as $\beta^t_i=\mathcal{V}^t_{i}/\sum_j^M \mathcal{V}^t_{j}$.
In the second training phase, $\beta^t_i$ is regarded as the weight of the $i$-th source subnet $\mathcal{G}^{S_i}$.
Therefore, the pseudo subnet parameters $P^t_{\mathcal{G}^{T}}$ at the $t$-th training step are updated as EMA of aggregated parameters of each $\mathcal{G}^{S_i}$:
\begin{small}
\begin{equation}
\begin{aligned}
\label{PSL}
P^t_{\mathcal{G}^T} = \alpha P^{t-1}_{\mathcal{G}^{T}} + (1-\alpha) \cdot \sum_i^M \beta_i^t P^{t}_{\mathcal{G}^{S_i}},
\end{aligned}
\end{equation}%
\end{small}
where $P^t_{\mathcal{G}^{S_i}}$ denotes the parameters of $\mathcal{G}^{S_i}$ at training step $t$ and $\alpha$ is a smoothing coefficient parameter.

\subsection{Consistency Learning of RPN}
Although we have learned a specialized subset for each source in the first training phase, we employ an implicit strategy to bridge the discrepancy between each source and the target in the second phase.
Considering that the performance of RPN is closely related to the detection results, we conduct a consistency regularization on the region proposals from $G^{S_i}_{p}$ and $G^{T}_{p}$ for the same target image, where $G^{S_i}_{p}$ and $G^{T}_{p}$ denote the RPN of $\mathcal{G}^{S_i}$ and $\mathcal{G}^{T}$, respectively.
For a target image, it will be fed into different subnets after $G_1$.
Then each RPN will generate top-$\mathcal{N}$ ranked proposed bounding boxes for detection.
We expect that the region proposals from different RPNs are consistent through each source subnet which has learned specialized domain knowledge.
We use $\mathcal{R}^i=\{r^i_j\}_{j=1}^{\mathcal{N}}$ to represent region proposal set of the $i$-th source subnet.
Similarly, the region proposal set of $G^{T}_p$ is denoted as $\mathcal{R}^T=\{r^T_j\}_{j=1}^{\mathcal{N}}$.
Since $\mathcal{G}^T$ is updated through aggregating the knowledge from labeled sources with DW strategy, it will progressively be more friendly for the target domain during training.
Therefore, we impose consistency regularization between each $\mathcal{R}^i$ and $\mathcal{R}^T$.
For the $n$-th region proposal $r^T_n \in \mathcal{R}^T$, the highest Intersection-over-Union (IoU) overlap between $\{r^i_j\}_{j=1}^{\mathcal{N}}$ and $r^T_n$ is denoted as $\mathcal{O}^i_n=\max_{j\in[1, \mathcal{N}]}({\rm IoU}(r^i_j,r^T_n))$, and the corresponding serial number of region proposal in $\mathcal{R}^i$ is $n^*_i=\arg\max_{j\in[1, \mathcal{N}]}({\rm IoU}(r^i_j,r^T_n))$.
The overall consistency cost is formalized as:
\begin{small}
\begin{equation}
\begin{aligned}
\label{con}
\mathcal{L}_{con}= \frac{1}{\mathcal{N}} \sum_{n=1}^\mathcal{N} \sum_{i=1}^\mathcal{N} \mathcal{O}^i_n |n^*_i-n|,
\end{aligned}
\end{equation}%
\end{small}
where $|n^*_i-n|$ represents the distance between $n^*_i$ and $n$.
\subsection{Overall Objective}
The supervised learning loss for the detection of labeled source images is denoted as $\mathcal{L}_{det}$, which is composed of classification and regression error for RPN and RCNN.
Combining detection loss $\mathcal{L}_{det}$ and our introduced losses for domain adaptation, the final loss function of DMSN is written as:
\begin{small}
\begin{equation}
\begin{aligned}
\label{lambda}
\mathcal{L}= \mathcal{L}_{det} + \lambda (\mathcal{L}_l + \mathcal{L}_h + \mathcal{L}_{con}),
\end{aligned}
\end{equation}%
\end{small}
where $\lambda$ is a trade-off parameter to balance the detection loss and domain adaptation loss.
Placing GRL between backbone and domain discriminator, $\mathcal{L}_l$ and $\mathcal{L}_h$ are optimized by an adversarial manner.
While domain discriminators aim to minimize $\mathcal{L}_l$ and $\mathcal{L}_h$, the base network optimizes its parameters to maximize the losses due to reversed gradient.
Note that $\mathcal{L}_{con}$ and PSL can automatically participate in the second training phase without manual work, and DMSN is optimized in an end-to-end training manner.
No gradients are propagated through pseudo subnet $\mathcal{G}^T$ in the training.
The parameters of $\mathcal{G}^T$ is updated via Equation~(\ref{PSL}).

%

\section{Experiments}
In evaluation, we conduct extensive experiments of our framework in different domain adaptation scenarios, including cross camera adaptation and cross time adaptation.
More results are included in the supplementary material.
%

\subsection{Datasets}
\noindent \textbf{KITTI.} The KITTI dataset~\cite{geiger2012we} is collected from an autonomous driving platform, containing street scenarios taken in cities, highways, and rural areas.
A total of 7,481 training images are used as source images.

\noindent \textbf{Cityscapes.} The Cityscapes dataset~\cite{cordts2016cityscapes} contains driving scenarios taken in cities.
There are 2,975 images in the training set and 500 images in the validation set with pixel-level labels.
All of them are transformed into bounding boxes annotations.

\noindent \textbf{BDD100k.} BDD100k dataset~\cite{yu2018bdd100k} contains 100k images, including 70k training images and 10k validation images with bounding boxes annotations.
The images are taken in the different times of day including \textit{daytime}, \textit{night}, and \textit{dawn/dusk}.
The images taken in \textit{daytime} as the target domain in cross camera adaptation, including 36,728 training and 5,258 validation images.
Besides, we use the images taken in all the times in cross time adaptation.
%

~\subsection{Compared Baselines}
In this paper, we compare DMSN with the following baselines:
(1) \textbf{Source-only}, \ie, train on the source images and test on the target images directly, indicating the lower bound of domain adaptation.
The original Faster R-CNN is treated as the detection model.
(2) \textbf{Single-source DA}, conduct MSDA via single-source DA, including Strong-Weak~\cite{saito2019strong}, SCL~\cite{shen2019scl}, DA-ICR-CCR~\cite{xu2020exploring}, SW-ICR-CCR~\cite{xu2020exploring}, GPA~\cite{xu2020cross}.
(3) \textbf{Multi-source DA}, extend existing MSDA methods for classification,\ie MDAN~\cite{zhao2018adversarial} and M$^3$SDA~\cite{peng2019moment}, to perform MSDA for object detection by utilizing these algorithms on feature extractor.
Note that all the DA methods in the paper are unsupervised, where the labels of target training set are unavailable during training.
Besides, we also report oracle results by directly training Faster R-CNN using training target data.
For the source-only and single-source domain adaptation, we apply two strategies:
(1) single-source, \ie conduct adaptation on each single source domain;
(2) source-combined, \ie directly combining all the sources into a unified domain.

\subsection{Implementation Details}
In our experiments, Faster R-CNN~\cite{ren2015faster} is employed as based object detector in which the pretrained VGG16~\cite{simonyan2014very} is adopted as the backbone, following~\cite{chen2018domain, xu2020exploring,saito2019strong}.
Unless otherwise stated, the shorter side of the image has a length of 600 pixels following the implementation of~\cite{ren2015faster, saito2019strong} with ROI-alignment~\cite{he2017mask}.
The learning rate is initialized as 0.001, and all the models are trained for 20 epochs with cosine learning rate decay.
At the 10-th epoch, PSL and consistency regularization begin to be performed until the end of the training.
We set $\gamma$ of Equation~(\ref{gamma}) as 5.0 and $\lambda$ of Equation~(\ref{lambda}) as 1.0  following~\cite{saito2019strong}, while $\alpha$ of Equation~(\ref{PSL}) and $\mathcal{N}$ of Equation~(\ref{con}) are set to 0.99 and 256, respectively.
In the testing phase, the test images of the target domain will be fed into all subnets after $G_1$.
The final predictions obtained by integrating results from different subnets.
We implemented all the experiments with Pytorch~\cite{paszke2017automatic}.
\begin{table}[t]
	\centering
	\footnotesize
	\setlength\tabcolsep{5pt}
	\caption{Experimental results on adaptation from Cityscapes and KITTI to BDD100k (\textit{daytime}). Average precision (AP, \%) on \textit{car} category in target domain is evaluated. The best result is highlight with \textbf{bold}. }
	\begin{tabular}{l| l| c }
		\toprule
		\toprule
		Standards    &  Methods      &     AP on car        \\ \midrule
		\multirow{3}{*}{Source-only} & Cityscapes          &    44.6               \\
		& KITTI              &    28.6                \\
		& Cityscapes+KITTI             &    43.2                 \\ \midrule
		\multirow{4}{*}{Cityscapes-only DA}&  Strong-Weak~\cite{saito2019strong}           &    45.5                \\
		&       SCL~\cite{shen2019scl}            &    46.3                \\
		&      DA-ICR-CCR~\cite{xu2020exploring}       &   45.3                   \\
		&      SW-ICR-CCR~\cite{xu2020exploring}              &        46.5            \\  \midrule
		\multirow{4}{*}{KITTI-only DA}&      Strong-Weak~\cite{saito2019strong}          &    29.6                 \\
		&       SCL~\cite{shen2019scl}            &    31.1                  \\
		&      DA-ICR-CCR~\cite{xu2020exploring}       &   29.2                   \\
		&      SW-ICR-CCR~\cite{xu2020exploring}             &      30.8               \\ \midrule
		\multirow{4}{*}{Source-combined DA}& Strong-Weak~\cite{saito2019strong}            &    41.9                  \\
		&       SCL~\cite{shen2019scl}            &    43.0                 \\
		&      DA-ICR-CCR~\cite{xu2020exploring}       &           41.3           \\
		&      SW-ICR-CCR~\cite{xu2020exploring}              &    43.6
		\\ \midrule
		\multirow{3}{*}{Multi-source DA}&  MDAN~\cite{zhao2018adversarial}             &          43.2           \\
		&     M$^3$SDA~\cite{peng2019moment}         &          44.1            \\
		&     DMSN (Ours)             &        \textbf{49.2}	
		\\  \midrule
		Oracle                       &     Faster R-CNN~\cite{ren2015faster}      &   60.2                 \\
		\bottomrule
		\bottomrule
	\end{tabular}
	\label{bdd100k_day}
\end{table}

\begin{table*}[t]
	\centering
	\footnotesize
	\setlength\tabcolsep{6pt}
	\caption{Adaptation between different subsets with in BDD100k dataset. \textit{daytime} and \textit{night} are the sources, while \textit{dawn/dusk} is the target domain. The mean average precision (mAP, \%) of 10 categories is evaluated.The best result of each category is highlight with \textbf{bold}.}
	\begin{tabular}{l| l | c  c  c  c c c c c c c | c}
		\toprule
		\toprule
		Standards    &  Methods       &   bike     &  bus      &  car      &    motor     &    person   &    rider    &    light & sign & train & truck & mAP               \\ \midrule
		\multirow{3}{*}{Source-only} & Daytime  &  35.1        & 51.7  & 52.6   &  9.9    & 31.9   &  17.8   & 21.6 & 36.3     & 0     & 47.1     &  30.4     \\
		& Night   &  27.9      & 32.5    &  49.4   & 15.0     & 28.7   & 21.8     & 14.0    &  30.5    &  0    & 30.7     &  25.0 \\
		&     Daytime+Night               &  31.5    & 46.9  & 52.9 & 8.4  & 29.5  & 21.6 & 21.7  & 34.3     &  0    &  42.2    &  28.9        \\ \midrule
		\multirow{5}{*}{Daytime-only DA}&   Strong-Weak~\cite{saito2019strong}    & 34.9 & 51.2 & 52.7 & 15.1 & 32.8 & 23.6  & 21.6 & 35.6 &  0 & 47.1 & 31.4  \\
		&           SCL~\cite{shen2019scl}        &  29.1 & 51.3 & 52.8 & 17.2 & 32.0 & 19.1 & 21.8 & 36.3      &  0    & 47.2  & 30.7       \\
		&      GPA~\cite{xu2020cross}       &  36.6      & 52.1  & 53.1   & 15.6     & 33.0   & 23.0    &  21.7   &  35.4   & 0    & 48.0     & 31.8    \\
		&      DA-ICR-CCR~\cite{xu2020exploring}              &   35.6        &  47.5   &  52.7   &  13.9    & 32.2   &  22.7    &  22.8    &  35.5    & 0     &  45.7    & 30.9    \\
		&      SW-ICR-CCR~\cite{xu2020exploring}       &   32.8     & 51.4  & 53.0   & 15.4     & 32.5   & 22.3    & 21.2    &   35.4  & 0    &  47.9    &   31.2  \\  \midrule
		\multirow{5}{*}{Night-only DA}&  Strong-Weak~\cite{saito2019strong}    &   31.4  & 38.2 & 51.0 & 9.9 & 29.5 & 22.2 & 18.7     &  32.5 &    0  & 35.7 & 26.9     \\
		&      SCL~\cite{shen2019scl}            &   25.3 & 31.7 & 49.3 & 8.9 & 25.8 & 21.2 & 15.0 & 28.6  & 0 & 26.2 & 23.2   \\
		&      GPA~\cite{xu2020cross}       &    32.7   &  38.3  & 51.8   & 14.1   & 29.0   & 21.5    & 17.1    & 31.1    & 0    & 40.0     &  27.6   \\
		&      DA-ICR-CCR~\cite{xu2020exploring}        &   30.0        & 32.4    &  50.1   & 14.4     & 29.1   &  22.8    &   17.4   & 32.2    & 0     &  29.7    & 25.8    \\
		&      SW-ICR-CCR~\cite{xu2020exploring}              & 32.3        &  45.1   &  51.6   & 7.2     & 29.2   &   24.9   &  19.9    &  33.0    &    0  & 41.1     &  28.4     \\      \midrule
		\multirow{5}{*}{Source-combined DA}&  Strong-Weak~\cite{saito2019strong}     &  29.7  & 50.0    & 52.9  & 11.0  & 31.4 & 21.1 &   23.3 & 35.1     & 0     &  44.9  & 29.9  \\
		&             SCL~\cite{shen2019scl}     & 33.9 & 47.8 & 52.5 & 14.0 & 31.4 & 23.8 & 22.3 & 35.4 &  0 & 45.1 & 30.9  \\
		&      GPA~\cite{xu2020cross}       &    31.7   & 48.8 & 53.9   & \textbf{20.8}     & 32.0   & 21.6    &  20.5   & 33.7    &  0   &  43.1    & 30.6    \\
		&      DA-ICR-CCR~\cite{xu2020exploring}              &   28.2     & 47.6    &  51.6   &  17.6    & 28.8   & 21.9     & 17.4     & 33.2     & 0     &  45.8    & 29.2    \\
		&      SW-ICR-CCR~\cite{xu2020exploring}              &  25.3        &  51.3   & 52.1    & 17.0     &  33.4  &  18.9    & 20.7     & 34.8     &  0    &  47.9    & 30.2      \\  \midrule
		\multirow{3}{*}{Multi-source DA}&  MDAN~\cite{zhao2018adversarial}  &   \textbf{37.1}   & 29.9  & 52.8    &  15.8    & 35.1   &  21.6    &  24.7   &  38.8    & 0     & 20.1     &  27.6     \\
		&         M$^3$SDA~\cite{peng2019moment}          &    36.9
		& 25.9    &  51.9   &  15.1    & 35.7   &  20.5    & 24.7  &  38.1    & 0     &  15.9    & 26.5  \\
		&        DMSN (Ours)          &36.5 & \textbf{54.3} & \textbf{55.5} & 20.4 & \textbf{36.9} & \textbf{27.7}& \textbf{26.4} & \textbf{41.6}  & 0     & \textbf{50.8} & \textbf{35.0} \\ \midrule
		Oracle&      Faster R-CNN~\cite{ren2015faster}     &   27.2   & 39.6    & 51.9    &  12.7    & 29.0   & 15.2     & 20.0  &   33.1   & 0     &  37.5    & 26.6  \\
		\bottomrule
		\bottomrule
	\end{tabular}
	\label{cross_time}
\end{table*}

~\subsection{Cross Camera Adaptation}
Different datasets are captured by different devices or setups, which exhibit different scenes and viewpoints, incurring a strong domain shift.
In this experiment, we show the adaptation between datasets collected by different cameras.
Specifically, KITTI and Cityscapes are used as two source domains, while the subset taken in the daytime of BDD100k is treated as target domain.
The experiment is an adaptation from small-scale datasets to a large-scale dataset.	
%

In Table~\ref{bdd100k_day}, we report experimental results evaluated on the \textit{car} category (common category) in terms of the average precision (AP).
In source-only, the results of KITTI and Cityscapes+KITTI are worse than that of Cityscapes, where ``Cityscapes+KITTI" means directly combining Cityscapes and KITTI.
%
%
The performance drop after adding KITTI can mainly attribute to two  reasons:
(1) the domain discrepancy between two sources is not well bridged.
(2) compared with Cityscapres, the domain distribution of KITTI is further with that of BDD100k, and simply combining KITTI with Cityscapes will decrease the performance.
%
Meanwhile, regardless of source-only or single-source DA methods, simply combining images of multiple sources to train a model results in the worse result compared with the best result trained by single domain.
Particularly, the Strong-Weak model transferred from cityscapes achieves 45.5\% AP, while the source-combined drops to 41.1\% AP.
It is mainly because that the domain discrepancy also exists in different sources, and thus the source-combined images may interfere with each other during training~\cite{riemer2018learning}.
Compared with existing multi-source DA methods~\cite{zhao2018adversarial, peng2019moment} for other tasks, our DMSN for object detection obtains obviously better performance.
It is observed that there still exists a performance gap between our method and oracle results.
In the future, more excellent unsupervised DA methods are required to bridge the gap.
\begin{table*}[t]
	\centering
	\footnotesize
	\setlength
	\tabcolsep{6pt}
	\caption{Ablation study for adaptation from \textit{daytime} and \textit{night} to \textit{dawn/dusk} within BDD100k dataset. SS denotes the source subnets and PS denotes the pseudo subnet. DW represents dynamic weighting strategy. The best result of each category is highlight with \textbf{bold}.}
	\begin{tabular}{c c c  c c c | c  c  c  c  c  c  c  c  c  c  |  c}
		\toprule
		\toprule
        SS& $\mathcal{L}_l$ &  $\mathcal{L}_h$ & PS & DW & $\mathcal{L}_{con}$ & bike & bus & car & motor & person & rider & light & sign & train & truck & mAP              \\ \midrule
		${\surd}$ &    &     &      &    &   & 31.4       &  23.0    &  52.3  & 18.4    & 35.3    &  21.4     &   24.5   &   39.2    &  0    &  35.0    & 28.0    \\
		${\surd}$ &  ${\surd}$        &       &             &               &           &   32.9       &  33.4   &  53.2    &  18.2       & 36.0     & 23.8    & 27.1     &  40.2    & 0     & 34.5    & 29.9     \\
		${\surd}$ &  ${\surd}$        &   ${\surd}$           &      &                &           & 30.4         &  52.7   & 54.7     &  12.2       & 34.5     &  22.6   &  24.9    &  38.0    &0      & 46.9 & 31.7 \\
		${\surd}$ &  ${\surd}$        &   ${\surd}$     &     ${\surd}$        &   &    &           33.0         & 47.4         &  54.8   &   14.5   &   36.4      &  26.0    &  \textbf{27.6}   &   41.4   &   0   &   47.8   &   32.9     \\
		${\surd}$ &  ${\surd}$        &   ${\surd}$     &    ${\surd}$   &     &  ${\surd}$        &    33.1       & \textbf{54.3}         & \textbf{55.6}   & 18.5     &  36.2       & 24.8     & 27.1    &  41.1    &   0   &  50.4    & 34.2   \\
		${\surd}$ &  ${\surd}$        &   ${\surd}$     &    ${\surd}$          &  ${\surd}$         &  ${\surd}$  &\textbf{36.5} & \textbf{54.3} & 55.5
		& \textbf{20.4} & \textbf{36.9} & \textbf{27.7} & 26.4 & \textbf{41.6}  & 0     & \textbf{50.8} & \textbf{35.0}      \\
		\bottomrule
		\bottomrule
	\end{tabular}
	\label{ablation}
\end{table*}
\begin{figure*}[t]
	\begin{center}
		\includegraphics[width=0.9\linewidth]{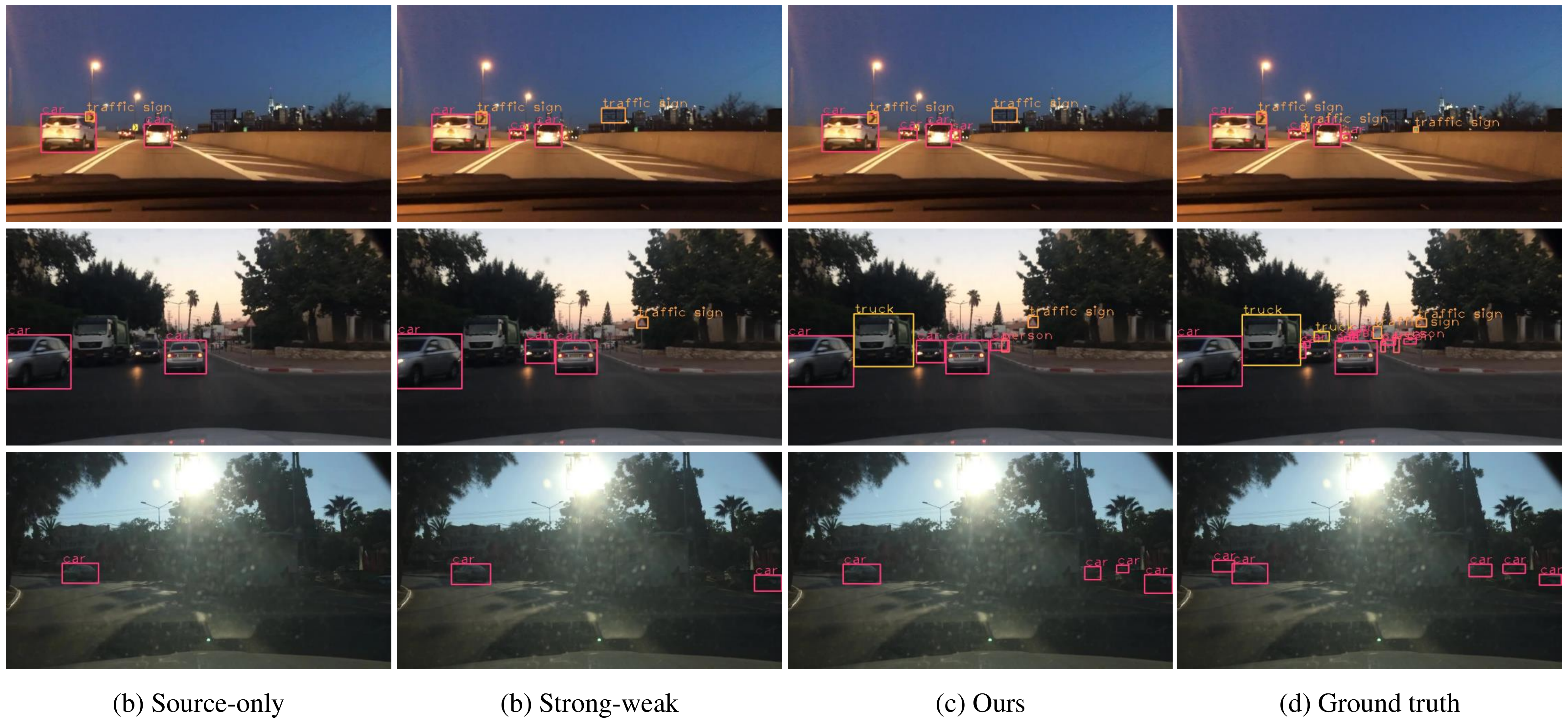}
		\caption{Examples of the qualitative results for adaptation from \textit{daytime} and \textit{night} subsets to \textit{dawn/dusk}. From left to right column, the results of source-only, source-combined DA via Strong-Weak~\cite{saito2019strong}, our framework DMSN, and the ground truth. }
		\label{fig: visualization}
	\end{center}
\end{figure*}

\subsection{Cross Time Adaptation}
In this scenario, we conduct our domain adaptation between different time conditions.
Specifically, we conduct experiments using three different time-related subsets of BDD100k, \ie \textit{daytime}, \textit{night}, \textit{dawn/dusk}. 36,728 training images in \textit{daytime} and 27,971 training images in \textit{night} are treated as source data, while 5,027 training images in \textit{dawn/dusk} are treated as target data.
The evaluation is conducted on a total of 778 validation dawn/dusk images.
Different from cross camera adaptation, this adaption scenario is from the large-scale datasets to a small-scale dataset.

Table~\ref{cross_time} shows that our model effectively reduces the domain discrepancy across time conditions and performs favorably against the state-of-the-art methods~\cite{saito2019strong, shen2019scl, xu2020exploring}.
Intuitively, when \textit{night} is treated as the source domain, the adaptation performance is worse than \textit{daytime} as the source.
It matches human cognition that it is hard to recognize and localize the object at night.
Due to the domain gap between \textit{daytime} and \textit{night}, the performance of source-combined DA drops compared with daytime-only DA apart from SCL~\cite{shen2019scl}.
Our adaptation model achieves considerable improvement over the best results of source-combined DA and multi-source DA by 4.1\% mAP and 7.4\% mAP, respectively.
%
%
Though multi-source DA (\ie, MDAN and M$^3$SDA) employ different methods to align different domains, the alignment is limited on the image level.
However, instance-level information plays an essential role in object detection.
Only using forcible image-level alignment may reduce the discriminative ability of local features for objects.
It is observed that the results of MDAN and M$^3$SDA are extremely poor on \textit{bus} and
\textit{truck} that have similar appearance, and it directly results in performance drop on mAP.
It is worth emphasizing that our results surpasses that of oracle by about 10\% mAP, which reveals a phenomenon that the oracle results may be not the upper bound of domain adaptation in all the scenarios.
We conclude that the above phenomenon is closely related to the number of training data.
In our experiments, a total of over 60,000 images of \textit{daytime} and \textit{night} participate in training model, which greatly exceeds 5,027 \textit{dawn/dusk} training images in the oracle experiment.
Based on the viewpoint, the contribution of our algorithm is more significant if we collect more labeled data from multiple source domains.

\vspace{-0.5mm}
\subsection{Analysis}
\noindent \textbf{Ablation study.}
We perform a detailed ablation study for cross time adaptation, and its results are shown in Table~\ref{ablation}.
Based on the skeleton of DMSN, the baseline is achieved by only conducting supervised learning for object detection in all the source subnets (SS), without any other operations.
When consecutively adding domain discriminator loss on the low-level and high-level features, the performance is improved by 1.9\% mAP and 1.8\%, respectively.
In the fourth row, our DW strategy cannot be applied for PSL without consistency regularization, so we treat each source domain equally when updating parameters of the pseudo subnet.
The results demonstrate the effectiveness of the pseudo subnet (PS).
It is worth highlighting that the newly designed instance-level consistency regularization for region proposals has obvious effectiveness, \ie a 1.3\% mAP gain.
Finally, the DW strategy for weighting parameters of different source subnets also contributes a slight performance gain.

\noindent \textbf{Visualization.}
In Figure~\ref{fig: visualization}, we present three example results from source-only, compared method Strong-Weak and ours in the cross time adaptation scenario.
In general, the proposed algorithm has merits in three aspects compared with existing state-of-the-art methods.
First, the proposed algorithm accurately detects the small object, \ie \textit{car} in the distance in the first example.
In the second example, our model correctly localizes the object, \ie \textit{truck}, that has less sample in training data.
Finally, in the situation of a strong backlight, \ie
the third example, the results demonstrate that our model can also well tackle the hard scenario.
Therefore, it is illustrated that our domain adaptation model has a robust generalization ability for these hard cases.

\section{Conclusion}
In this paper, we develop a framework, \ie DMSN, to address a new task that is adaptation from multiple source domains for object detection.
We propose hierarchical feature alignment for features of different levels.
Meanwhile, we intend to preserve the characteristics of each source via supervised learning in an independent subnet.
We also introduce a pseudo subnet that is updated by exponential moving average parameters in different source subnets.
The final prediction is derived from the aggregation of multiple subnets.
The results on different adaptation scenarios demonstrate the effectiveness of our method.
We also reveal that as long as enough source images are collected, it is possible that adaptation results surpass the oracle ones.

\section*{Acknowledgment}
This work was supported by the National Key Research and Development Program of China Grant (No. 2018AAA0100403), the National Natural Science Foundation of China (Nos. 61876094, U1933114, 61701273), Natural Science Foundation of Tianjin, China (Nos. 20JCJQJC00020, 18JCYBJC15400, 18ZXZNGX00110).

{\small
\bibliographystyle{ieee_fullname}
\bibliography{egbib}
}

\end{document}